\PassOptionsToPackage{table,dvipsnames}{xcolor}

\documentclass[conference]{IEEEtran}
\IEEEoverridecommandlockouts
\usepackage{cite}
\usepackage{amsmath,amssymb,amsfonts}
\usepackage{algorithmic}
\usepackage{graphicx}
\usepackage{textcomp}
\usepackage{xcolor}
\usepackage{booktabs}
\usepackage{multirow}
\usepackage{graphicx}
\usepackage{subcaption} 
\usepackage{soul}

\usepackage[colorinlistoftodos,prependcaption,textsize=tiny]{todonotes}
\newcommand{\comment}[1]{}
\usepackage{comment}
\usepackage{subcaption}
\usepackage{colortbl} 
\usepackage{array}          
\usepackage{tabularx}       
\usepackage[hidelinks]{hyperref}

\usepackage[font=footnotesize]{caption}

\newcommand\copyrighttext{%
  \footnotesize \textcopyright 2025 IEEE. Personal use of this material is permitted.  Permission from IEEE must be obtained for all other uses, in any current or future media, including reprinting/republishing this material for advertising or promotional purposes, creating new collective works, for resale or redistribution to servers or lists, or reuse of any copyrighted component of this work in other works.
  }
\newcommand\copyrightnotice{%
\begin{tikzpicture}[remember picture,overlay]
\node[anchor=south,yshift=10pt] at (current page.south) {\fbox{\parbox{\dimexpr\textwidth-\fboxsep-\fboxrule\relax}{\copyrighttext}}};
\end{tikzpicture}%
}

\def\BibTeX{{\rm B\kern-.05em{\sc i\kern-.025em b}\kern-.08em
    T\kern-.1667em\lower.7ex\hbox{E}\kern-.125emX}}
\begin{document}

\title{Physics-Informed Machine Learning for Vessel Shaft Power and Fuel Consumption Prediction: Interpretable KAN-based Approach\\ 
}

\author{
\IEEEauthorblockN{Hamza Haruna Mohammed}
\IEEEauthorblockA{\textit{Simula Research Laboratory} \\
Oslo, Norway\\
hamzahm@simula.no}
\and
\IEEEauthorblockN{Dusica Marijan}
\IEEEauthorblockA{\textit{Simula Research Laboratory} \\
Oslo, Norway\\
dusica@simula.no}
\and
\IEEEauthorblockN{Arnbjørn Maressa}
\IEEEauthorblockA{\textit{Navtor AS} \\
Egersund, Norway\\
arnbjorn.maressa@navtor.com}
}

\maketitle
\copyrightnotice

\begin{abstract}
Accurate prediction of shaft rotational speed, shaft power, and fuel consumption is crucial for enhancing operational efficiency and sustainability in maritime transportation. Conventional physics-based models provide interpretability but struggle with real-world variability, while purely data-driven approaches achieve accuracy at the expense of physical plausibility. This paper introduces a Physics-Informed Kolmogorov–Arnold Network (PI-KAN), a hybrid method that integrates interpretable univariate feature transformations with a 
physics-informed loss function and a leakage-free chained prediction pipeline. Using operational and environmental data from five cargo vessels, PI-KAN consistently outperforms the traditional polynomial method and neural network baselines. The model achieves the lowest mean absolute error (MAE) and root mean squared error (RMSE), and the highest coefficient of determination (R²) for shaft power and fuel consumption across all vessels, while maintaining physically consistent behavior. Interpretability analysis reveals rediscovery of domain-consistent dependencies, such as cubic-like speed–power relationships and cosine-like wave and wind effects. These results demonstrate that PI-KAN achieves both predictive accuracy and interpretability, offering a robust tool for vessel performance monitoring and decision support in operational settings.

\end{abstract}


 \section{Introduction}\label{introduction}

Accurate prediction of shaft RPM, shaft power, and fuel consumption is critical for optimizing the operational efficiency of maritime vessels. Traditional approaches rely on either physics-based model derived from propeller theory and engine thermodynamics \cite{SaettoneTavakoliTaskarJensenothers2020}, or purely data-driven
methods such as neural networks \cite{AssaniMatiKatelanavka2023}. While Physics-based models provide interpretability and adherence to fundamental principles, they often struggle to capture the complex, real-world variability observed in vessel operations. Conversely, data-driven models excel at learning patterns from historical data but may produce physically implausible predictions due to their black-box nature \cite{LiaoKttig2016}.

Recent hybrid methods attempt to bridge this gap by combining physical knowledge with machine learning. Some approaches integrate physical equations as constraints within neural networks \cite{BourchasPapalambrou2025}, while others employ linear combinations of physics-based and data-driven components \cite{LiaoKttig2016}. However, these methods often struggle to balance the influence of physical laws with empirical data, leading to either overly rigid or unconstrained predictions. Moreover, interpretability, a key requirement for maritime engineers remains limited in many deep learning-based solutions
\cite{ZikasGkirtzouFilippopoulosKalatzisothers2025}.

We propose Physics-informed Kolmogorov-Arnold Networks (PI-KAN), a novel hybrid method that addresses these limitations through three key innovations. First, it employs per-feature univariate transformations followed by a linear combination, enabling flexible yet interpretable modeling. Unlike conventional neural networks \cite{liu2024kan}, PI-KAN decomposes the prediction task into simpler, explainable components while retaining the capacity to learn nonlinear relationships. Second, a physics-informed loss function dynamically balances data fidelity with physical consistency using an auto-tuned parameter $\gamma$. This ensures that predictions adhere to fundamental propulsion principles without sacrificing accuracy. Third, the method incorporates a chained training pipeline with out-of-fold stacking to prevent target leakage across sequential predictions (RPM → power → fuel consumption). Vessel-wide tuning of $\gamma$ further enhances generalization across diverse vessel types.

The primary contributions of this work are:
\begin{itemize}
    \item \textbf{Interpretable Hybrid Architecture}: PI-KAN combines univariate feature transformations with linear combinations, offering transparency in how input features influence predictions. This directly addresses the interpretability requirements in maritime applications (Sections~\ref{interpretability-requirements-in-maritime-applications} and \ref{interpretable-and-vessel-wide-modeling}), distinguishing it from opaque black-box deep learning models.\\
    
    \item \textbf{Adaptive Physics Data Balancing}: The auto-tuned $\gamma$ parameter ensures that physical constraints are neither overbearing nor negligible, adapting to the predictive uncertainty of each vessel. This mechanism outperforms fixed-weight physics-informed models by reducing bias in adverse conditions and maintaining accuracy across vessels.\\

    \item \textbf{Leakage-Free Sequential Prediction}: The chained pipeline with out-of-fold stacking prevents information leakage, enabling robust multi-target modeling. Compared to conventional multi-task neural networks, this approach yields lower error propagation between shaft rpm, power, and fuel consumption, improving vessel-wide generalisation.\\

\end{itemize}

The remainder of this paper is organized as follows: Section \ref{related-work} reviews related work in vessel performance modeling and hybrid approaches. Section \ref{background-on-kolmogorovarnold-networks-and-marine-performance-modelling} introduces the Kolmogorov-Arnold Network (KAN) framework and its adaptation for maritime systems. Section \ref{methodology-physics-Informed-kan-for-vessel-fuel-and-power-prediction} details the PI-KAN architecture, physics-informed loss function, and training pipeline. Similarly, the same Section \ref{methodology-physics-Informed-kan-for-vessel-fuel-and-power-prediction} describes the experimental setup and datasets in subsection \ref{dataset}. Section \ref{results-and-comparative-analysis} presents comparative results, and Section \ref{discussion-and-conclusion} discusses vessel-wide generalization and future directions.

\section{Related Work}\label{related-work}

Vessel performance modeling has undergone significant evolution in recent years, driven by the growing availability of sensor data and the increasing demand for energy-efficient operations. Existing approaches can be broadly categorized into physics-based models, data-driven methods, and hybrid techniques that integrate both paradigms.

\subsection{Physics-Based Modeling}\label{physics-based-modeling}

Traditional physics-based approaches rely on first-principles equations derived from propeller hydrodynamics and engine thermodynamics. These models, such as those based on propeller cube law and torque-speed relationships, provide interpretable predictions but often struggle with real-world variability due to simplifications in their formulations \cite{Karagiannidis2019}. For instance, the influence of hull fouling or weather conditions is typically approximated using empirical correction factors, which may not generalize across different vessel types or operational conditions.

Recent advancements have sought to refine these models by incorporating high-fidelity computational fluid dynamics (CFD) simulations \cite{KanZhengChenZhouDaiBinamaothers2020}. While such methods improve accuracy, they remain computationally expensive and impractical for real-time vessel-scale predictions.

\subsection{Data-Driven Approaches}\label{data-driven-approaches}

Machine learning techniques have gained traction for their ability to capture complex patterns from historical operational data. Methods like random forests and gradient boosting have been applied to predict fuel consumption and shaft power \cite{KrishnappaKamangarBaigKhanothers2023}. Deep learning models, particularly recurrent neural networks (RNNs), have also been explored for time-series forecasting of vessel performance metrics \cite{AbebeShinNohLeeLee2020}.

However, purely data-driven models often lack interpretability and may produce physically implausible predictions when extrapolating beyond training conditions. For example, a neural network might predict decreasing fuel consumption with increasing speed, a violation of basic
propulsion principles \cite{KaragiannidisThemelis2021}.

\subsection{Hybrid Methods}\label{hybrid-methods}

To address these limitations, hybrid methods have emerged as a promising direction. Physics-informed neural networks (PINNs) incorporate physical constraints directly into the loss function, ensuring predictions adhere to governing equations \cite{BourchasPapalambrou2025}. Another approach involves blending physics-based submodels with data-driven corrections, as seen in \cite{LiangHanVanemothers2025}, where a baseline physical model is augmented with a neural network to capture residual errors.

Despite their advantages, existing hybrid methods face challenges in balancing the influence of physics and data. Fixed weighting schemes often lead to either overly rigid predictions or insufficient physical consistency. Moreover, many approaches lack interpretability, making it
difficult for maritime engineers to validate model behavior.

\subsection{Interpretable and Vessel-Wide
Modeling}\label{interpretable-and-vessel-wide-modeling}

Interpretability has become a key requirement for maritime applications, where engineers need to understand and trust model predictions. Techniques like SHAP (SHapley Additive exPlanations) have been applied to post-hoc explain black-box models \cite{ZikasGkirtzouFilippopoulosKalatzisothers2025}. However, these methods add computational overhead and may not fully reveal the underlying decision logic.

Vessel-wide generalization is another significant challenge, as models must adapt to the diverse types of vessels and varying operating conditions. Some studies have explored transfer learning or meta-learning to improve cross-vessel performance \cite{KalafatelisNomikosGiannopoulosothers2024}. Nevertheless, these methods often require extensive retraining or fail to maintain physical plausibility across the vessel.

\subsection{Comparison with Proposed
Method}\label{comparison-with-proposed-method}

The proposed PI-KAN method distinguishes itself from existing approaches in several key aspects. Unlike fixed hybrid models, it dynamically balances physics and data via an auto-tuned loss-weighting parameter, ensuring adaptability across different operational regimes. The use of univariate transformations and linear combinations provides inherent interpretability, allowing engineers to inspect feature contributions directly. Furthermore, the chained prediction pipeline with out-of-fold stacking prevents leakage while maintaining physical consistency. These innovations enable PI-KAN to achieve superior accuracy and generalizability compared to conventional physics-based, data-driven, or hybrid methods.

\begin{figure*}[!t]
\centering
\vspace{-2mm}
\includegraphics[width=0.90\textwidth]{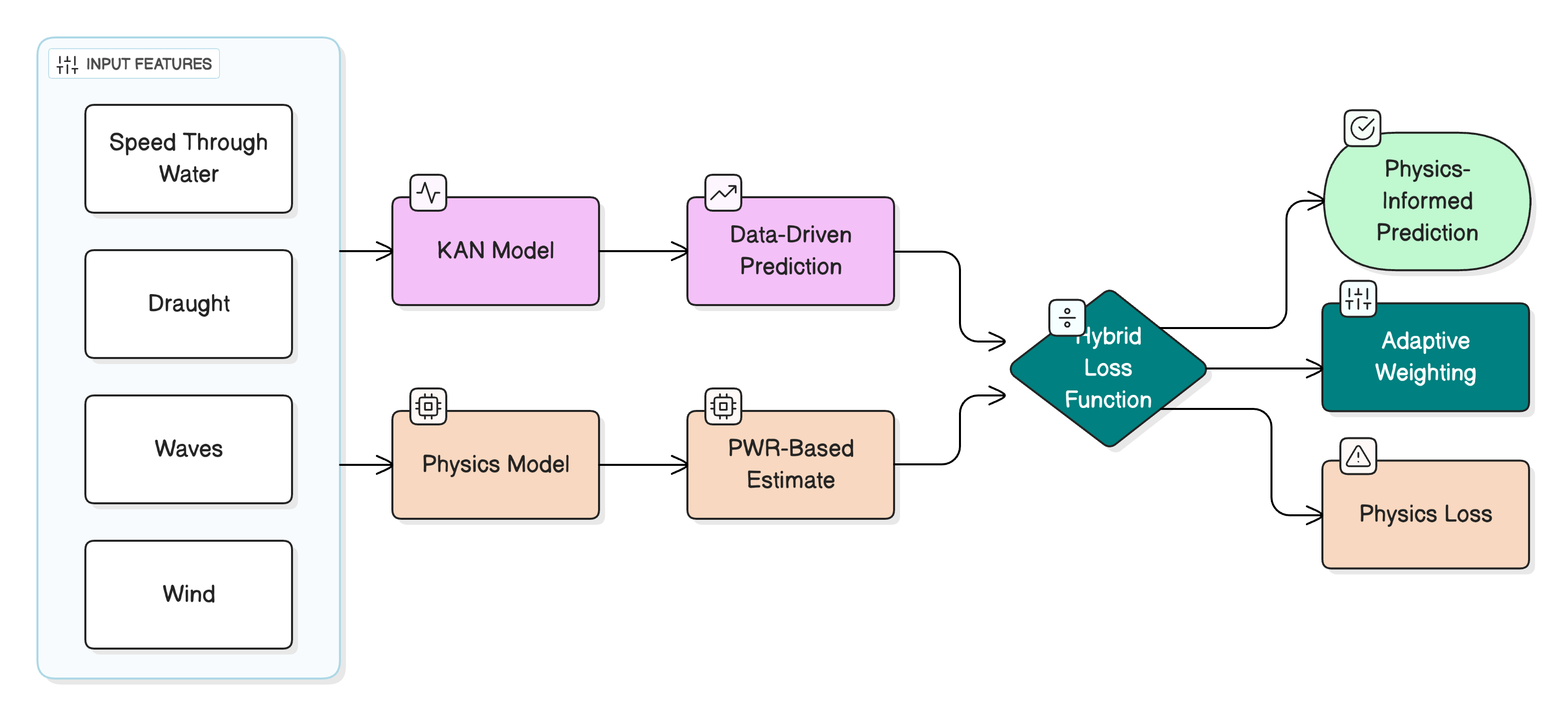}
\vspace{-2mm}
\caption{
Architecture of the Physics-Informed Kolmogorov–Arnold Network (PI-KAN) for vessel performance modeling. 
The model applies per-feature univariate transformations followed by a linear combination layer, incorporates a physics-informed loss for propulsion consistency, and uses a leakage-free chained pipeline (RPM $\rightarrow$ shaft power $\rightarrow$ fuel consumption) with out-of-fold stacking to ensure physically plausible and robust multi-stage predictions as described in the Figure \ref{fig:architecture-chained-prediction-vessel-performnace}.
}
\label{fig:architecture-physics-Informed-kan}
\vspace{-3mm}
\end{figure*}

\section{Background on Kolmogorov--Arnold Networks and Vessel Performance Modelling}\label{background-on-kolmogorovarnold-networks-and-marine-performance-modelling}

To establish the theoretical foundation for our proposed method, we first examine the mathematical framework of Kolmogorov-Arnold Networks (KANs) and their relevance to vessel performance modelling. The Kolmogorov-Arnold representation theorem, a fundamental result in approximation theory, states that any multivariate continuous function can be expressed as a finite composition of univariate functions and additions \cite{Liu2015} \cite{liu2024kan}. This theoretical insight provides the basis for KAN architectures, which differ fundamentally from conventional neural networks in their structural composition.

\subsection{Mathematical Foundations of
KANs}\label{mathematical-foundations-of-kans}

The Kolmogorov-Arnold representation decomposes a multivariate function \(f: [0,1]^d \rightarrow \mathbb{R}\) into a sum of univariate functions:

\[f(x_1, ..., x_d) = \sum_{q=1}^{2d+1} \Phi_q \left( \sum_{p=1}^d \phi_{p,q}(x_p) \right) \tag{1}\]

where \(\phi_{p,q}\) and \(\Phi_q\) are continuous univariate functions. This decomposition suggests that complex multivariate relationships can be broken down into simpler, interpretable components, a property particularly valuable for vessel systems, where physical interpretability is crucial. Unlike traditional multilayer perceptrons that apply nonlinear transformations to multivariate inputs, KANs implement this theorem by learning univariate functions along edges of a computational graph \cite{Hornik1991}.

\subsection{KANs vs.~Traditional Neural
Networks}\label{kans-vs.-traditional-neural-networks}

Three key differences distinguish KANs from conventional neural network architectures. First, while standard networks use fixed activation functions (e.g., ReLU, sigmoid) applied to weighted sums of inputs, KANs learn adaptive univariate functions for each input feature. Second, the additive structure of KANs naturally lends itself to interpretability, as the contribution of each component function can be examined independently. Third, the network width (i.e., the number of univariate functions) grows linearly with the input dimension, rather than exponentially, making KANs more parameter-efficient for high-dimensional problems.
\cite{QianLiuLiuWuWong2018}.

\subsection{Vessel Performance Modelling
Challenges}\label{marine-performance-modelling-challenges}

Vessel propulsion systems exhibit complex nonlinear behaviors that challenge traditional modelling approaches. The relationship between shaft rotational speed (RPM) and power output, for instance, follows a roughly cubic trend under ideal conditions, but actual measurements deviate due to factors like hull fouling, weather conditions, and engine wear \cite{TerzievTezdoganIncecik2022}. These real-world complexities create a tension between physical first principles and data-driven corrections - precisely the scenario where KANs' hybrid nature offers advantages.

Physical models based on propeller theory typically express shaft power \(P\) as:

\[P = 2\pi n Q \tag{2}\]

where \(n\) is shaft speed and \(Q\) is torque. However, this idealized relationship overlooks the numerous environmental and operational factors that influence actual vessel performance. Data-driven approaches can capture these effects, but often lose connection to the underlying physics. The KAN framework provides a middle ground, where physical relationships can be encoded as priors while still allowing data-driven refinement of individual components.

\subsection{Historical Development of Vessel Performance
Models}\label{historical-development-of-marine-performance-models}

The evolution of vessel performance models has progressed through three distinct generations. First-generation models relied entirely on physical equations derived from naval architecture principles \cite{Falzarano2018}. Second-generation models introduced empirical corrections to account for observed deviations from theory \cite{Holtrop1977}. The current third generation combines these approaches through machine learning while attempting to preserve
interpretability \cite{LangWuMao2022}.

KANs represent a natural progression in this evolution, as their structure aligns with the way maritime engineers traditionally decompose propulsion systems into subsystems (engine, propeller, hull) with well-characterized individual behaviors. The network's additive
structure mirrors the common practice of analyzing power losses separately for different components before combining them into a complete system model \cite{NaujoksStedenMullerothers2007}.

\subsection{Interpretability Requirements in Maritime
Applications}\label{interpretability-requirements-in-maritime-applications}

Unlike many machine learning applications where prediction accuracy alone suffices, maritime engineering demands models that provide actionable insights. Regulatory compliance, maintenance planning, and operational optimization all require understanding not just what the
model predicts, but why \cite{AbuellaAtouiNowaczykJohanssonothers2023}.
KANs address this need through their transparent architecture, each univariate function corresponds to a specific input feature's transformation, and the final prediction combines these transformed inputs linearly.

This interpretability proves particularly valuable when analyzing anomalous vessel behavior. For instance, if a particular ship's fuel consumption deviates from vessel norms, engineers can examine the individual feature transformations to identify whether the discrepancy
stems from engine performance, hull condition, or operational patterns \cite{CoradduOnetoGhioSavioothers2016}. Such diagnostic capability remains challenging with conventional neural networks, where feature interactions are complex and opaque.

\section{Methodology: Physics-Informed KAN for Vessel Fuel and Power Prediction}\label{methodology-physics-Informed-kan-for-vessel-fuel-and-power-prediction}

The Physics-Informed KAN (PI-KAN) framework introduces a novel hybrid architecture that combines the interpretability of univariate feature transformations with the predictive power of physics-informed learning. The method addresses three critical challenges in maritime performance modeling:

\begin{itemize}
    \item maintaining physical plausibility while learning from data,
    \item preventing information leakage in sequential predictions, and 
    \item ensuring consistent performance across diverse vessel types.
\end{itemize}

\begin{figure}[!t]
\centering
\vspace{-2mm}
\includegraphics[width=1.01\columnwidth]{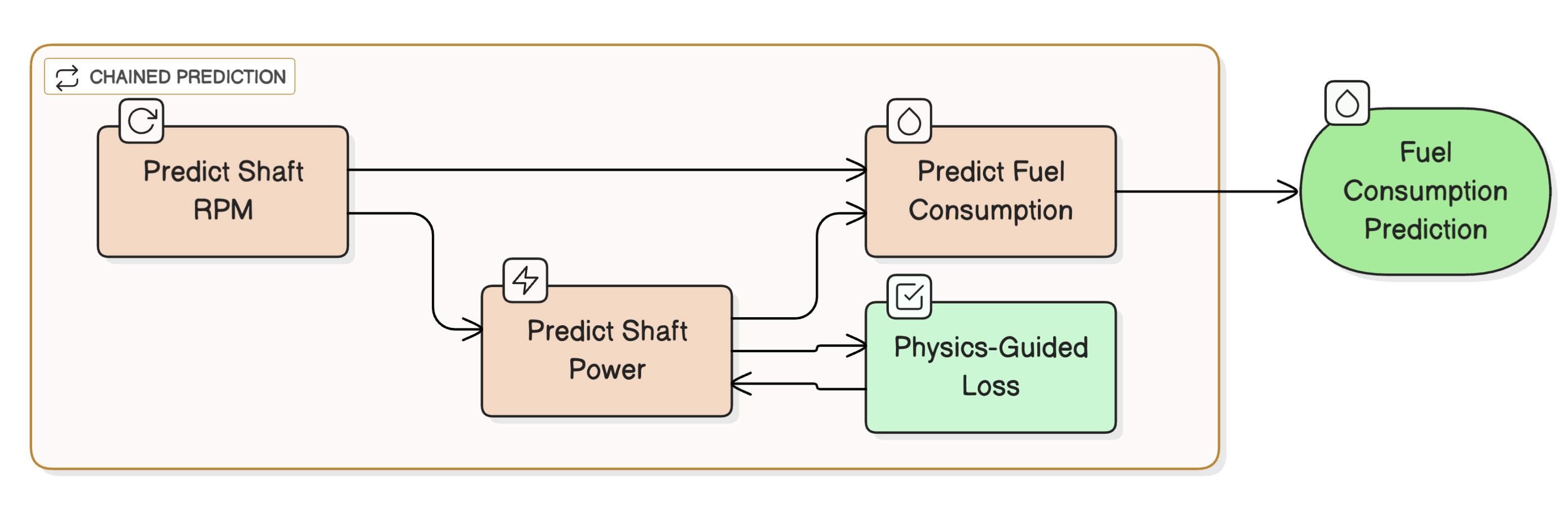}
\vspace{-2mm}
\caption{
Leakage-free chained prediction pipeline used in PI-KAN: shaft RPM is predicted first, followed by shaft power and then fuel consumption, with each stage receiving out-of-fold (OOF) predictions from the previous one to avoid target leakage and preserve physical causality.
}
\label{fig:architecture-chained-prediction-vessel-performnace}
\vspace{-3mm}
\end{figure}

\begingroup
\setlength{\tabcolsep}{4pt}
\renewcommand{\arraystretch}{1.25}
\captionsetup{font=footnotesize} 
\footnotesize
\begin{table}[ht]
    \centering
    \caption{Categorized input features used for shaft rpm, shaft power, and fuel consumption prediction.}
    \renewcommand{\arraystretch}{1.25}
    \rowcolors{2}{gray!10}{white}
    \begin{tabularx}{\linewidth}{p{2.5cm} >{\raggedright\arraybackslash}p{1.5cm} >{\arraybackslash}X}
        \toprule
        \textbf{Category} & \textbf{Feature} & \textbf{Description} \\
        \midrule

        \rowcolor{white}\multicolumn{3}{l}{\textbf{Operational}}\\
        & $V$ & Speed through water (knots) \\
        & $T$ & Vessel draught (m), distance from waterline to keel \\
        \addlinespace[2pt]

        \rowcolor{white}\multicolumn{3}{l}{\textbf{Environmental}}\\
        & $depth_\text{sea}$ & Sea depth below keel (m) \\
        & $t_\text{sea}$ & Sea surface temperature (°C) \\
        & $h_\text{wave}$ & Significant wave height (m) \\
        & $T_\text{wave}$ & Wave peak period (s) \\
        \addlinespace[2pt]

        \rowcolor{white}\multicolumn{3}{l}{\textbf{Swell}}\\
        & $h_\text{swell}$ & Swell height (m) \\
        & $T_\text{swell}$ & Swell period (s) \\
        & $d_\text{swell}$ & Swell direction relative to heading (°) \\
        & $d_\text{wave}$ & Wave direction relative to heading (°) \\
        \addlinespace[2pt]

        \rowcolor{white}\multicolumn{3}{l}{\textbf{Wind}}\\
        & $v_\text{wind}$ & Apparent wind speed (m/s) \\
        & $d_\text{wind}$ & Apparent wind direction (° relative to vessel) \\
        \addlinespace[2pt]

        \rowcolor{white}\multicolumn{3}{l}{\textbf{Derived proxies}}\\
        & $V^3$ & Propeller cube-law proxy (cubed vessel speed) \\
        & $\cos(d_\text{wave})$ & Directional proxy for wave incidence \\
        \addlinespace[2pt]

        \rowcolor{white}\multicolumn{3}{l}{\textbf{Stacked (model outputs)}}\\
        & $P_{rpm}$ & Predicted shaft RPM → (input to power model) \\
        & $P_{shaft\_power}$ & Predicted shaft power →  (input to fuel model) \\
        \bottomrule
    \end{tabularx}
    \label{tab:features}
\end{table}
\endgroup

Overall, the dataset provides a diverse and realistic foundation for evaluating vessel performance-prediction methods across various operational conditions, vessel sizes, and maintenance states. Figure~\ref{fig:architecture-physics-Informed-kan}  and Figure~\ref{fig:architecture-chained-prediction-vessel-performnace} illustrate the general framework of the proposed two-stage prediction approach.  

\subsection{Architecture of PI-KAN for Vessel Performance Modeling}\label{architecture-of-PI-kan-for-marine-systems}

The core of PI-KAN consists of per-feature univariate transformations implemented as small multilayer perceptrons (MLPs), followed by a linear combination layer. For an input feature vector \(\mathbf{x} = [x_1, ..., x_d]\), each feature \(x_p\) undergoes a nonlinear transformation through a dedicated MLP \(\phi_p\):

\[h_p = \phi_p(x_p) \tag{3}\]

Where \(\phi_p\) is a two-layer neural network with sigmoid activation functions. The transformed features \(h_p\) are then combined linearly to produce the final prediction \(\hat{y}\):

\[\hat{y} = \sum_{p=1}^d w_p h_p + b \tag{4}\]

Here, \(w_p\) and \(b\) are learnable weights and bias, respectively. This architecture ensures that each feature's contribution to the prediction remains interpretable, as the influence of \(x_p\) on \(\hat{y}\) is mediated solely through its corresponding transformation \(\phi_p\).

In vessel performance prediction, the input features typically include shaft RPM, vessel speed, draft, and environmental conditions (e.g., wind speed, wave height). The output targets are shaft power \(P\) and fuel consumption \(\dot{m}_f\), predicted sequentially to maintain physical causality.

\begingroup
\setlength{\tabcolsep}{4pt}
\renewcommand{\arraystretch}{1.25}
\captionsetup{font=footnotesize} 
\footnotesize
\begin{table*}[ht]
    \centering
    \caption{Number of instances and data periods for each vessel in training and test sets, including dry docking dates.}
    \renewcommand{\arraystretch}{1.5}
    \begin{tabular}{|p{1.0cm}|p{2.4cm}|p{2.2cm}|p{3.9cm}|p{3.9cm}|p{2.2cm}|}
        \rowcolor{gray!20}
        \hline
        \textbf{Vessel} & \textbf{Train Set Instances} & \textbf{Test Set Instances} & \textbf{Train Dates (Start → End)} & \textbf{Test Dates (Start → End)} & \textbf{Dry Docking Date} \\
        \hline
        \textbf{A-6} & 12,758 & 21,930 &  2023-01-09 → 2024-01-15 & 2024-01-15 → 2025-09-01 &  14/12/2022 \\
        \hline
        \textbf{A-8} & 18,145 & 8,663 & 2020-04-18 → 2020-10-13 & 2020-10-18 → 2021-09-14 & 14/04/2022 \\
        \hline
        \textbf{A-10} & 20,608 & 91,664 & 2020-05-11 → 2022-03-27 & 2022-04-22 → 2025-03-29 & 23/05/2023 \\
        \hline
        \textbf{A-12} & 23,186 & 96,258 & 2020-09-21 → 2023-02-19 & 2023-02-22 → 2025-05-07 & 09/11/2023 \\
        \hline
        \textbf{A-13} & 22,970 & 70,077 & 2020-04-20 → 2021-11-20 & 2021-11-22 → 2025-05-02 & 14/08/2023 \\        
        \hline
    \end{tabular}
    \label{tab:dataset}
\end{table*}
\endgroup

\subsection{Physics-Informed Modeling and Physics-Informed Loss Function}
\label{physics-Informed-loss-function}

The training objective combines data fidelity with physical consistency:
\[
\mathcal{L} \;=\; \mathcal{L}_{\text{data}} \;+\; \lambda\,\mathcal{L}_{\text{physics}} \;+\; \mathcal{L}_{\text{reg}} \tag{5}
\]
where \(\mathcal{L}_{\text{data}}\) is the mean absolute error (MAE) between predictions and observations, 
\(\mathcal{L}_{\text{physics}}\) enforces adherence to first principles, and \(\mathcal{L}_{\text{reg}}\) is an ElasticNet term to reduce overfitting.

\paragraph{Physics loss via empirical resistance.}
We compute the physically required shaft power \(P_{\text{Physical}}\) using the empirical resistance model for calm-water, wind, and wave resistance given by physics equations, and the power as follows: `
\[
P_{\text{Physical}} \;=\; \big(R_{\text{Calm}} + R_{\text{Wind}} + R_{\text{Wave}}\big)\,V \, 
\]
The physics term for the power head penalizes deviations from this EF-based power:
\[
\mathcal{L}_{\text{PWR}} \;=\; \mathrm{MAE}\!\big(\,\hat{P},\, P_{\text{Physical}}\,\big) . \tag{6}
\]
For the fuel head, we enforce consistency with a thermal-efficiency relation:
\[
\mathcal{L}_{\text{FUEL}} \;=\; \mathrm{MAE}\!\big(\,\dot{\hat{m}}_{f},\, P/(\eta H)\,\big) , \tag{7}
\]
where \(\eta\) is the (effective) engine efficiency and \(H\) the fuel’s lower heating value.

In our implementation, the overall physics loss is
\[
\mathcal{L}_{\text{physics}} \;=\; \mathcal{L}_{\text{PWR}} \;+\; \mathcal{L}_{\text{FUEL}} \, .
\]

\paragraph{Optional cube-law prior with vessel-specific \(k\).}
As an auxiliary, we additionally penalize deviation from the propeller cube law,
\[
\tilde{\mathcal{L}}_{\text{cube}} \;=\; \mathrm{MAE}\!\big(\,\hat{P},\, k\,n^{3}\,\big),
\]
where \(n\) is shaft RPM and \(k\) is a \emph{vessel-specific} constant reflecting propeller and drivetrain characteristics. 
In practice, \(k\) is pre-calibrated per vessel from training data (e.g., \(k=\arg\min_{k}\,\mathrm{MAE}(P,kn^{3})\), or the robust statistic \(k=\mathrm{median}(P/n^{3})\) over valid segments). 
When used, we add this term with a small weight: \(\mathcal{L}_{\text{physics}}=\mathcal{L}_{\text{PWR}}+\gamma\,\tilde{\mathcal{L}}_{\text{cube}}+\mathcal{L}_{\text{FUEL}}\) (\(\gamma\!\in\![0,1]\)).

\paragraph{Auto-balancing of physics weight.}
The balancing parameter \(\lambda\) is updated during training to keep \(\mathcal{L}_{\text{data}}\) and \(\mathcal{L}_{\text{physics}}\) on comparable scales:
\[
\lambda_{\text{new}}
=\mathrm{clip}\!\Big(
\lambda_{\text{prev}}\,\exp\!\big(\eta\,(\mathcal{L}_{\text{data}}-\mathcal{L}_{\text{physics}})\big),\;
\lambda_{\min},\lambda_{\max}\Big) . \tag{8}
\]
This prevents either term from dominating and adapts the strength of physics guidance to the current prediction uncertainty.

\subsection{Chained Training with Out-of-Fold
Stacking}\label{chained-training-with-out-of-fold-stacking}

To predict shaft RPM, power, and fuel consumption sequentially without leakage, PI-KAN employs a chained pipeline with out-of-fold (OOF) stacking:

\begin{enumerate}
\def\labelenumi{\arabic{enumi}.}
\item
  \textbf{Shaft RPM Prediction}: Train \(f_{\text{RPM}}\) on folds
\(\{1,...,K-1\}\), predict on fold \(K\).\\
\item
  \textbf{Power Prediction}: Use OOF RPM predictions as input to train
  \(f_{\text{PWR}}\) on folds \(\{1,...,K-1\}\), predict on fold
  \(K\).\\
\item
  \textbf{Fuel Prediction}: Use OOF power predictions as input to train
  \(f_{\text{FUEL}}\).
\end{enumerate}

This approach ensures that downstream models (e.g., fuel prediction) never see the true values of upstream targets (e.g., RPM) during training, mimicking real-world deployment where future observations are unavailable.

\subsection{Vessel-Wide Hyperparameter
Tuning}\label{vessel-wide-hyperparameter-tuning}

The auto-balancing parameter \(\lambda\) is optimized across the entire vessel to ensure consistent physical adherence. For each vessel \(v\) in the vessel:

\begin{enumerate}
\def\labelenumi{\arabic{enumi}.}
\item
  Split data into training/validation sets
  \(\mathcal{D}_v^{\text{train}}, \mathcal{D}_v^{\text{val}}\).\\
\item
  Train PI-KAN on \(\mathcal{D}_v^{\text{train}}\) with candidate
  \(\lambda\) values.\\
\item
  Select \(\lambda\) that minimizes median MAE across all
  \(\mathcal{D}_v^{\text{val}}\).
\end{enumerate}

This vessel-wide tuning adapts the physics guidance to the collective behavior of vessels while preventing overfitting to individual ships' idiosyncrasies.

\subsection{Interpretability Through Univariate Responses}\label{interpretability-through-univariate-responses}

The per-feature transformations \(\phi_p\) provide direct insight into how input variables influence predictions. For example, the learned function \(\phi_{\text{RPM}}\) typically exhibits a near-cubic. The relationship for shaft power prediction aligns with propeller theory. Environmental features, such as wave height, exhibit threshold effects that become significant only when they exceed certain magnitudes.

These interpretable patterns allow maritime engineers to validate the model behavior against domain knowledge and identify anomalous vessel performance. For instance, an unexpectedly steep \(\phi_{\text{draft}}\) might indicate hull fouling, while a nonlinear \(\phi_{\text{wind}}\) could reveal windage effects specific to certain ship types.

The combination of these components, interpretable architecture, adaptive physics guidance, leakage-free training, and vessel-wide tuning, enables PI-KAN to achieve both high accuracy and physical plausibility. The method's modular design also facilitates the incorporation of additional physical constraints or domain-specific modifications as needed for particular applications.

\section{Experimental Evaluation}
\label{sec:experimental_evaluation}

\subsection{Compared Modeling Approaches}
\label{compared-modeling-approaches}

\paragraph{Polynomial Model.}
The PM is modeled using a multiplicative function of univariate polynomials:
\[
\text{PM}(u) = \prod_{i=1}^{n} p_i(u_i),
\]
where $n$ is the number of features, $u$ is the input vector, and $p_i$ is the polynomial function applied to the feature $u_i$.  Features are selected iteratively by examining correlation with residual error, resulting in four input features:  speed through water ($V$), draught ($T$), apparent wind speed ($v_\text{wind}$), and swell height ($h_\text{swell}$).  A third-order polynomial is used for each feature. This formulation is computationally lightweight and interpretable, performing competitively on shaft rpm, but it underfits more complex nonlinearities in shaft power and fuel consumption.

\paragraph{Multilayer Perceptron (MLP).}
As a data-driven baseline, we implement an MLP model whose input matches the selected (and engineered) features, with two hidden layers (37, 28; ReLU) and a linear head. Training uses Adam ($lr = 10^ (-2)$), a batch size of 8, and up to 200 epochs with early stopping. The objective is MSE plus a regularization penalty ($\alpha=0.01$, $l1\_ratio=0.5$). While flexible for nonlinear mappings, the MLP remains a black box with limited interpretability and sensitivity to hyperparameters/regularization.

\paragraph{PI-KAN} 
 Each KAN module applies per-feature univariate subnetworks (Linear–activation–LayerNorm stacks) concatenated and linearly combined, optimized with a physics-guided MAE loss that integrates elastic-net regularization $(\alpha{=}10^{-2},\rho{=}0.5)$ and a weighted physics term ($\lambda$ tuned or auto-balanced). We train with Adam (lr $10^{-3}$), batch size of $32$, MAE loss, and early stopping (up to 150 epochs) using an 80/20 split.
 
\subsubsection{Training} 
 Leakage-free learning is ensured by 5-fold out-of-fold stacking, producing \texttt{predicted\_rpm} and \texttt{predicted\_shaft\_power} for sequential targets. A chained pipeline (RPM $\rightarrow$ power $\rightarrow$ fuel) feeds predictions downstream, and fleet-wide $\lambda$ values are tuned by sweeping candidates and selecting the best median MAE across vessels (Figure~\ref{fig:architecture-chained-prediction-vessel-performnace}).

\subsection{Experimental Evaluation}
\label{experimental-evaluation}

To evaluate the effectiveness of PI-KAN, we consider the following research questions (RQs):

\begin{itemize}
    \item \textbf{RQ1:} Does PI-KAN improve prediction accuracy and physical plausibility compared to Polynomial, MLP, and standard KAN baselines?
    \item \textbf{RQ2:} Can PI-KAN generalize vessel-wide across ships with different operational and environmental conditions?
    \item \textbf{RQ3:} How interpretable are the learned transformations, and do they align with maritime domain knowledge (e.g., cubic speed–power law, cosine directional effects)?
\end{itemize}

We design our evaluation to explicitly address these questions. Subsequent subsections present the compared methods, datasets, and evaluation metrics.

\subsection{Dataset}\label{dataset}
We use 15\,min operational and environmental logs from five cargo vessels (A-6, A-8, A-10, A-12, A-13). Table~\ref{tab:dataset} reports train/test spans, sample counts, and most recent dry-dock dates. Coverage is heterogeneous: A-10/A-12 exceed 110k samples over more than 5 years; A-6 is balanced; A-8/A-13 are shorter and useful for assessing vessel-wide generalization. Maintenance records include scheduled dry-docking and mid-cycle propeller polishing; days-since-drydock were \emph{not} used as inputs but are analysed separately.

Features for performance modeling include operational/hydrodynamic variables (speed through water, $V$, RPM, draught, $T$, sea depth, and sea temperature) and environmental forcings (wave/swell height and direction, wind speed and direction; wind direction is relative to heading). Wave and wind fields are sourced from the Copernicus Marine Service~\cite{copernicus2025}. 

\subsection{Evaluation Metrics}\label{evaluation_metrics}
To assess the performance of the vessel performance prediction method, we employ four commonly used regression metrics:  

\begin{itemize}
    \item \textbf{Mean Absolute Error (MAE)}:  
    Measures the average absolute deviation between predicted values $\hat{y}_i$ and ground truth $y_i$, defined as:
    \begin{equation*}
        \text{MAE} = \frac{1}{n} \sum_{i=1}^{n} |y_i - \hat{y}_i|
    \end{equation*}
    where $n$ is the number of data points. MAE provides a straightforward measure of prediction accuracy.  

    \item \textbf{Root Mean Square Error (RMSE)}:  
    Similar to MAE but penalizes larger errors more heavily by squaring the deviations before averaging:
    \begin{equation*}
        \text{RMSE} = \sqrt{\frac{1}{n} \sum_{i=1}^{n} (y_i - \hat{y}_i)^2}
    \end{equation*}
    RMSE emphasizes the impact of significant outliers.  

    \item \textbf{Mean Absolute Percentage Error (MAPE)}:  
    Expresses prediction error as a percentage relative to the true values:
    \begin{equation*}
        \text{MAPE} = \frac{1}{n} \sum_{i=1}^{n} \left| \frac{y_i - \hat{y}_i}{y_i} \right| \times 100\%
    \end{equation*}
    providing an interpretable, scale-independent measure of accuracy.  

    \item \textbf{Coefficient of Determination ($R^2$)}:  
    Indicates the proportion of variance in the ground truth explained by the predictions:
    \begin{equation*}
        R^2 = 1 - \frac{\sum_{i=1}^{n} (y_i - \hat{y}_i)^2}{\sum_{i=1}^{n} (y_i - \bar{y})^2}
    \end{equation*}
    where $\bar{y}$ is the mean of the actual values. Higher $R^2$ values denote better model fit.  
\end{itemize}

\begin{figure}[t]
  \centering

  \begin{subfigure}[t]{\columnwidth}
    \centering
    \includegraphics[width=\columnwidth]{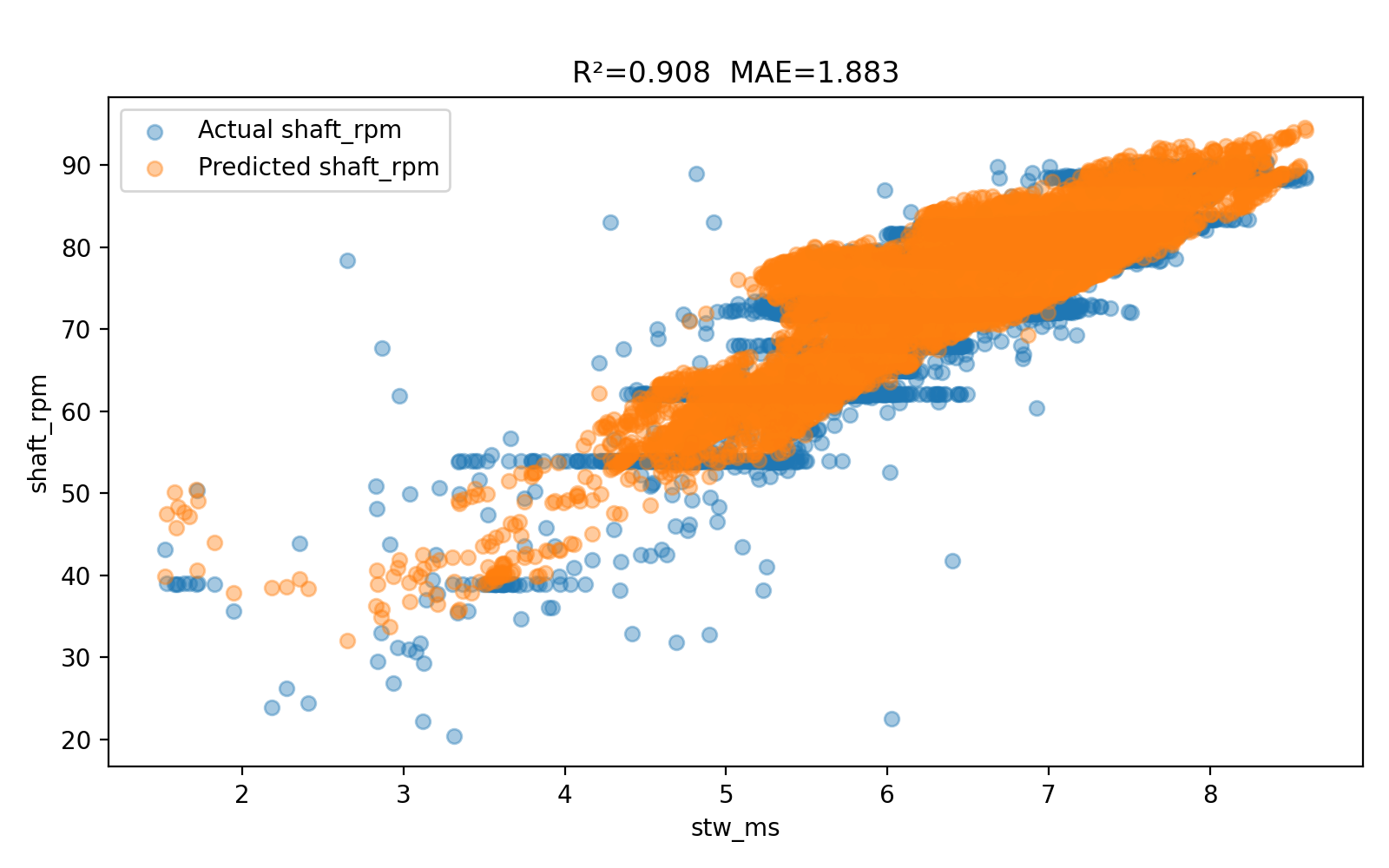}
    \caption{A-6 Vessel: Shaft RPM vs speed}
    \label{fig:rpm-vs-speed}
  \end{subfigure}

  \vspace{0.6em}

  \begin{subfigure}[t]{\columnwidth}
    \centering
    \includegraphics[width=\columnwidth]{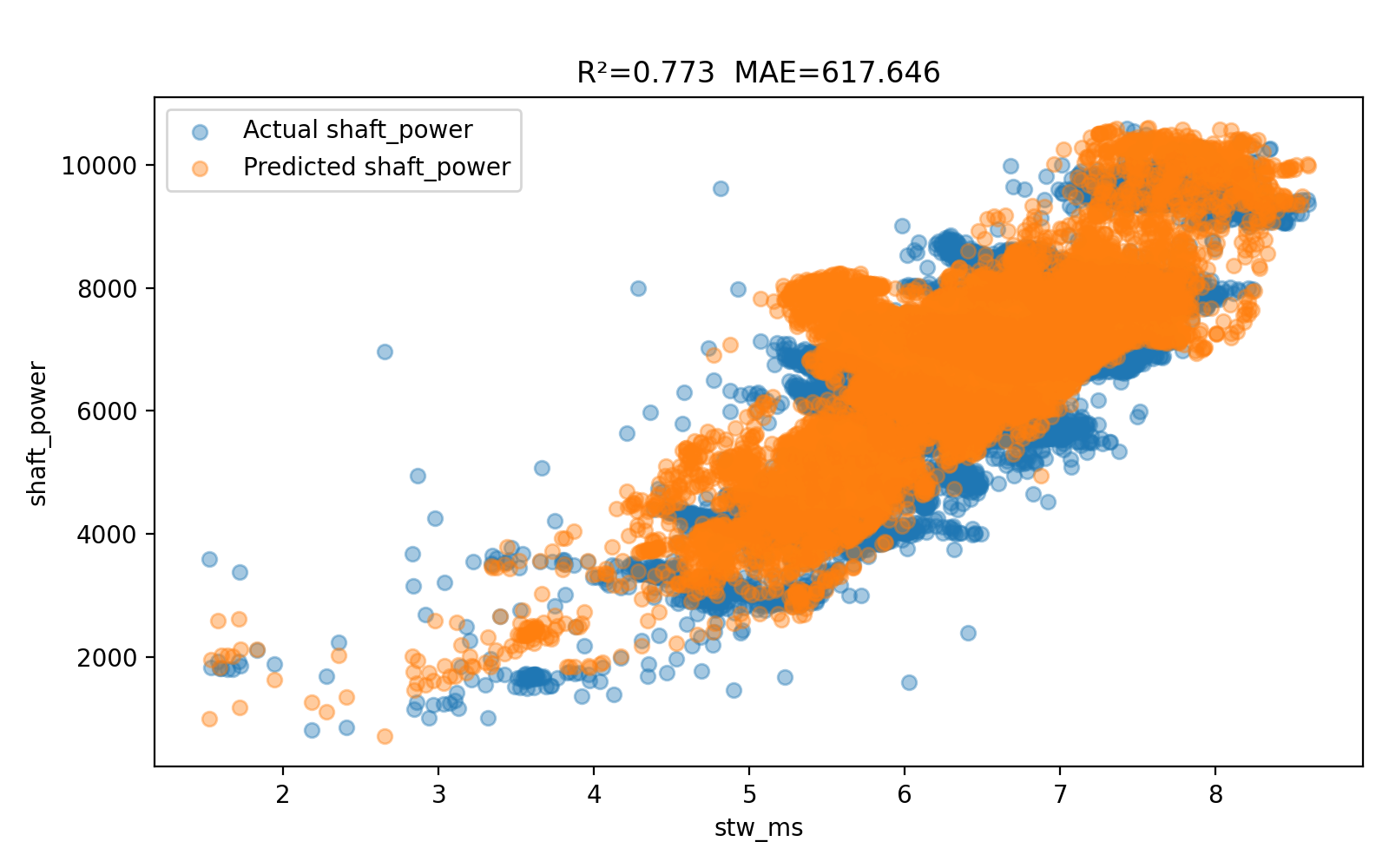}
    \caption{A-6 Vessel: Shaft power vs speed}
    \label{fig:power-vs-speed}
  \end{subfigure}

  \vspace{0.6em}

  \begin{subfigure}[t]{\columnwidth}
    \centering
    \includegraphics[width=\columnwidth]{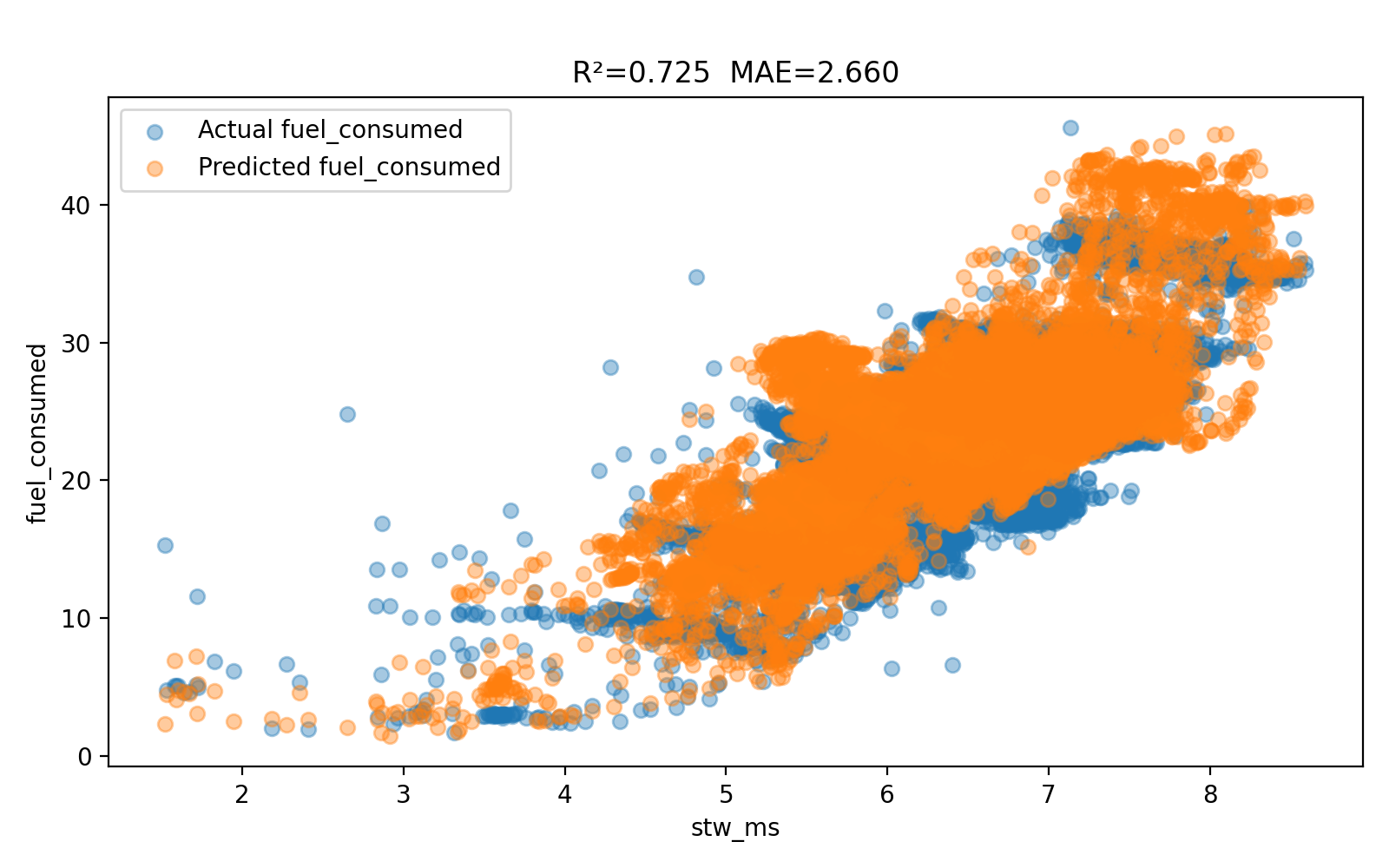}
    \caption{A-6 Vessel: Fuel consumed vs speed}
    \label{fig:fuel-vs-speed}
  \end{subfigure}

  \caption{Comparison of actual vs predicted performance metrics for A-6 Vessel.}
  \label{fig:vessel-performance}
\end{figure}

\section{Results and Comparative Analysis}\label{results-and-comparative-analysis}

\begingroup
\setlength{\tabcolsep}{4pt}
\renewcommand{\arraystretch}{1.25}
\captionsetup{font=footnotesize} 
\footnotesize
\begin{table*}[ht]
    \centering
    \caption{Benchmark performance on the test set (relative errors, overall values) across different targets. 
    \textbf{MAE:} Mean Absolute Error, \textbf{RMSE:} Root Mean Squared Error, \textbf{R$^2$:} Coefficient of Determination.}
    \renewcommand{\arraystretch}{2.0} 
    \rowcolors{2}{gray!10}{white}
    \begin{tabularx}{\textwidth}{
        p{2.0cm} p{2.2cm}
        *{3}{>{\centering\arraybackslash}X}
        *{3}{>{\centering\arraybackslash}X}
        *{3}{>{\centering\arraybackslash}X}
    }
        \toprule
        \rowcolor{gray!30}
        \textbf{Vessel} & \textbf{Method} &
        \multicolumn{3}{c}{\textbf{shaft\_rpm}} &
        \multicolumn{3}{c}{\textbf{shaft\_power}} &
        \multicolumn{3}{c}{\textbf{fuel\_consumed}} \\
        \cmidrule(lr){3-5} \cmidrule(lr){6-8} \cmidrule(lr){9-11}
        & &
        \textbf{MAE} & \textbf{RMSE} & \textbf{R$^2$} &
        \textbf{MAE} & \textbf{RMSE} & \textbf{R$^2$} &
        \textbf{MAE} & \textbf{RMSE} & \textbf{R$^2$} \\
        \midrule
        
        \multirow{4}{*}{A-6}
            & Polynomial   & 2.34  & \textbf{2.97}  & 88.37  & 12.39  & 15.80 & 60.34  & 17.52  & 22.98  & 2.01 \\
         \textbf{A-6}    & MLP          & 6.74  & 8.19  & 11.91  & 13.55  & 17.93  & 48.93  & 25.18  & 31.24  & -81.04 \\
            & KAN          & 10.42  & 11.61  & 77  & 33.52  & 40.47  & 61.51 & 26.01 & 25.25  & 71.42 \\
            & PI-KAN       & \textbf{1.88} & 3.33 & \textbf{90.79} & \textbf{8.89} & \textbf{10.96} & \textbf{77.27} & \textbf{11.24} & \textbf{13.98} & \textbf{72.47} \\
        \midrule
        
        \multirow{4}{*}{A-8}
            & Polynomial   & 2.34  & 3.02  & 85.85  & 13.98  & 18.18 & 34.08  & 16.42  & 21.05 & -7.89 \\
           \textbf{A-8}  & MLP          & 2.64 & 3.36 & 82.52 & 9.76 & 12.10 & 70.80 & 17.14 & 21.48 & -12.36 \\
            & KAN          & 3.17  & 3.90  & 76.34  & 15.24  & 18.64  & 30.70  & 12.44 & 15.27  & 43.20 \\
            & PI-KAN       & \textbf{1.91}  & \textbf{2.78}  & \textbf{86.61}  & \textbf{7.60}  & \textbf{9.49}  & \textbf{71.64}  & \textbf{9.31}  & \textbf{11.27}  & \textbf{66.09} \\
        \midrule
        
        \multirow{4}{*}{A-10}
            & Polynomial   & \textbf{2.51} & \textbf{3.76} & 12.69  & 21.29  & 23.42  & -17  & 25.18 & 26.93 & -39.7 \\
           \textbf{A-10}  & MLP          & 12.20  & 14.96  & -12.82  & 12.52  & 15.13  & -11.34  & 70.80  & 78.71  & -16.64 \\
            & KAN          & 7.73  & 8.94  & -39.4  & 28.14  & 31.70  & -40.2  & 13.59  & 17.08  & -63.3 \\
            & PI-KAN       & 3.04  & 4.22  & \textbf{48.86}  & \textbf{10.05}  & \textbf{14.32}  & \textbf{54.04}  & \textbf{9.10}  & \textbf{13.27}  & \textbf{15.46} \\
        \midrule
        
        \multirow{4}{*}{A-12}
            & Polynomial   & \textbf{3.14}  & \textbf{4.38}  & \textbf{80.61}  & 22.17  & 26.72  & 1.24  & 24.88  & 31.41  & -20.62 \\
           \textbf{A-12}  & MLP          & 9.52  & 10.64  & -14.55  & 29.72  & 31.78  & -39.64 & 12.17 & 15.91  & -47.87 \\
            & KAN          & 7.39  & 9.01  & 18.00  & 25.11  & 28.94  & -15.88  & 16.29  & 20.42  & 31.15 \\
            & PI-KAN       & 5.01  & 6.29  & 61.18  & \textbf{10.57}  & \textbf{15.14}  & \textbf{70.13}  & \textbf{9.54}  & \textbf{14.24}  & \textbf{68.93} \\
        \midrule
        
        \multirow{4}{*}{A-13}
            & Polynomial   & \textbf{2.35} & \textbf{3.26} & \textbf{73.24} & 16.48  & 18.52  & -13.55  & 25.58 & 28.66 & -26.90 \\
           \textbf{A-13}  & MLP          & 19.37  & 23.75  & -13.23  & 53.62  & 61.11  & -34.35  & 26.16  & 30.44  & -11.15 \\
            & KAN          & 15.02  & 17.08 & -63.66 & 41.64 & 36.88 & -47.43 & 32.81 & 38.42 & -56.3 \\
            & PI-KAN       & 5.22  & 5.93  & 54.16  & \textbf{10.33}  & \textbf{12.88}  & \textbf{57.16} & \textbf{15.60} & \textbf{18.16}  & \textbf{16.00} \\
        \bottomrule
    \end{tabularx}
    \label{tab:benchmark-overall}
\end{table*}
\endgroup

\begingroup
\setlength{\tabcolsep}{4pt}
\renewcommand{\arraystretch}{1.25}
\captionsetup{font=footnotesize} 
\footnotesize
\begin{table*}[ht]
    \centering
    \caption{Performance analysis of adversarial weather conditions (fuel\_consumption).\\
    Test voyage window = departure for a duration of travel in that period.  (All other voyages (different voyage\_ids) go into training). 
    Traditional model results are added for comparison.}
    \renewcommand{\arraystretch}{1.2}
    \rowcolors{2}{gray!10}{white}
    \begin{tabularx}{\textwidth}{l l c c *{2}{>{\centering\arraybackslash}X} *{2}{>{\centering\arraybackslash}X}}
        \toprule
        \textbf{Vessel} & \textbf{Departure (UTC)} & \textbf{Duration} & \textbf{Samples} 
        & \multicolumn{2}{c}{\textbf{PI-KAN}} & \multicolumn{2}{c}{\textbf{Traditional}} \\
        \cmidrule(lr){5-6} \cmidrule(lr){7-8}
        & & & & ME\% (signed) & MAPE\% & ME\% (signed) & MAPE\% \\
        \midrule
        A-1 & 2020-07-28 18:30:00 & 10 days 02:45 & 944 & 22.41 & 22.63 & 12.00 & -- \\
        A-11 & 2020-11-29 03:15:00 & 13 days 05:30 & 179 & 5.55 & 10.17 & -8.00 & -- \\
        \bottomrule
    \end{tabularx}
    \label{tab:voyage-summary}
\end{table*}
\endgroup

\subsection{Results for RQ1 and RQ2: Vessel-Wide Performance Evaluation}\label{vessel-wide-performance-evaluation}

Using the consolidated benchmark in Table~\ref{tab:benchmark-overall}, the Physics-Informed KAN (PI-KAN) attains the strongest
\emph{vessel-wide} performance on the \emph{energy-related targets}:
it achieves the lowest MAE and RMSE for \textbf{shaft\_power} and \textbf{fuel\_consumed} across \emph{all five} vessels (A-6, A-8, A-10, A-12, A-13), and simultaneously yields the highest $R^{2}$ in those targets for each vessel.
In contrast, for \textbf{shaft\_rpm} the \emph{Polynomial} baseline attains the lowest MAE/RMSE on 3/5 vessels (A-10, A-12, A-13), with PI-KAN leading on A-8 and a little bit on A-6 vessel. Notably, PI-KAN still secures the top $R^{2}$ on shaft\_rpm for 3/5 vessels (A-6, A-8, A-10), indicating better explained variance even where absolute-error minima are held by Polynomial.

These results align with expectations: power and fuel are tightly constrained by physics (e.g., propeller laws and energy balance), where physics guidance helps; rpm behaves more linearly and is well captured by a low-bias polynomial regressor as illustrated in Figure~\ref{fig:vessel-performance}, showing the speed versus target (rpm, shaft power, and fuel consumption) relationship.

\subsection{Results for RQ1: Target-Specific Analysis}\label{target-specific-analysis}

\paragraph{Shaft RPM.}
Polynomial regression yields the smallest absolute errors across most vessels, suggesting that rpm dynamics within the observed envelope are well approximated by low-order trends. However, PI-KAN often achieves higher $R^{2}$ values (A-6, A-8, A-10), suggesting improved variance explanation and greater robustness to distributional shifts.

\paragraph{Shaft Power.}
PI-KAN consistently achieves the lowest MAE/RMSE and the highest $R^{2}$ across all vessels, while preserving the expected cubic-like relationship between rpm and power. The physics-informed loss curbs implausible responses in high-power regimes and stabilizes learning under sparse, extreme conditions.

\paragraph{Fuel Consumption.}
PI-KAN is uniformly best (lowest MAE/RMSE, highest $R^{2}$) across all vessels. Improvements are most pronounced on vessels and periods with stronger environmental forcing (e.g., A-8, A-10), where physics constraints regularize the mapping and reduce error propagation from upstream targets.

\subsection{Results for RQ2: Adversarial Weather Case Study}\label{adversarial-weather-case-study}
\begin{figure}[!t]
\centering
\vspace{-2mm}
\includegraphics[width=0.51\textwidth]{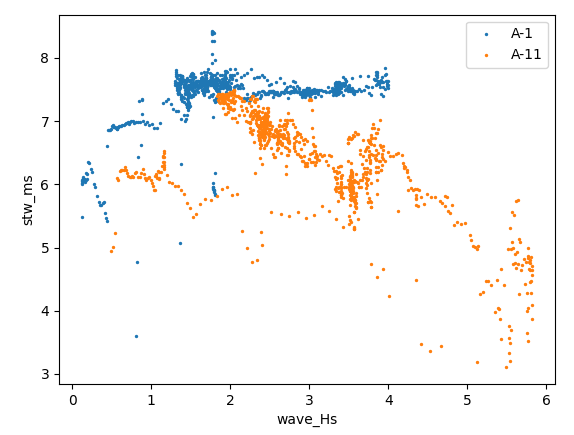}
\vspace{-2mm}
\caption{Adversarial weather response (A\textendash1 vs.\ A\textendash11): vessel speed as a function of wave height. 
A\textendash11 reduces speed at higher waves, while A\textendash1 holds a near-constant setpoint. 
This difference aligns with Table~\ref{tab:voyage-summary}: lower error on A\textendash11 (ME=+5.55\%, MAPE=10.17\%) versus higher error on A\textendash1 (ME=+22.41\%, MAPE=22.63\%).}
\label{fig:adversarial-weather-condition-result}
\vspace{-3mm}
\end{figure}

Under adverse conditions (Table~\ref{tab:voyage-summary}), PI\textendash KAN yields MAPE of 22.63\% (A\textendash1) and 10.17\% (A\textendash11), with signed ME of +22.41\% and +5.55\%, respectively. The traditional baseline shows signed ME of +12\% (A\textendash1) and $-8\%$ (A\textendash11), but MAPE was not computed, limiting direct comparison. Figure~\ref{fig:adversarial-weather-condition-result} shows that A\textendash11 reduces speed as waves increase, whereas A\textendash1 maintains a near-constant setpoint; the resulting variance in power/fuel explains the higher error on A\textendash1. Overall, PI-KAN is robust in aggregate, but voyage-level bias can increase under heavy seas, motivating policy-aware calibration for rough-weather operations.

\begin{figure}[!t]
\centering
\vspace{-2mm}
\includegraphics[width=0.51\textwidth]{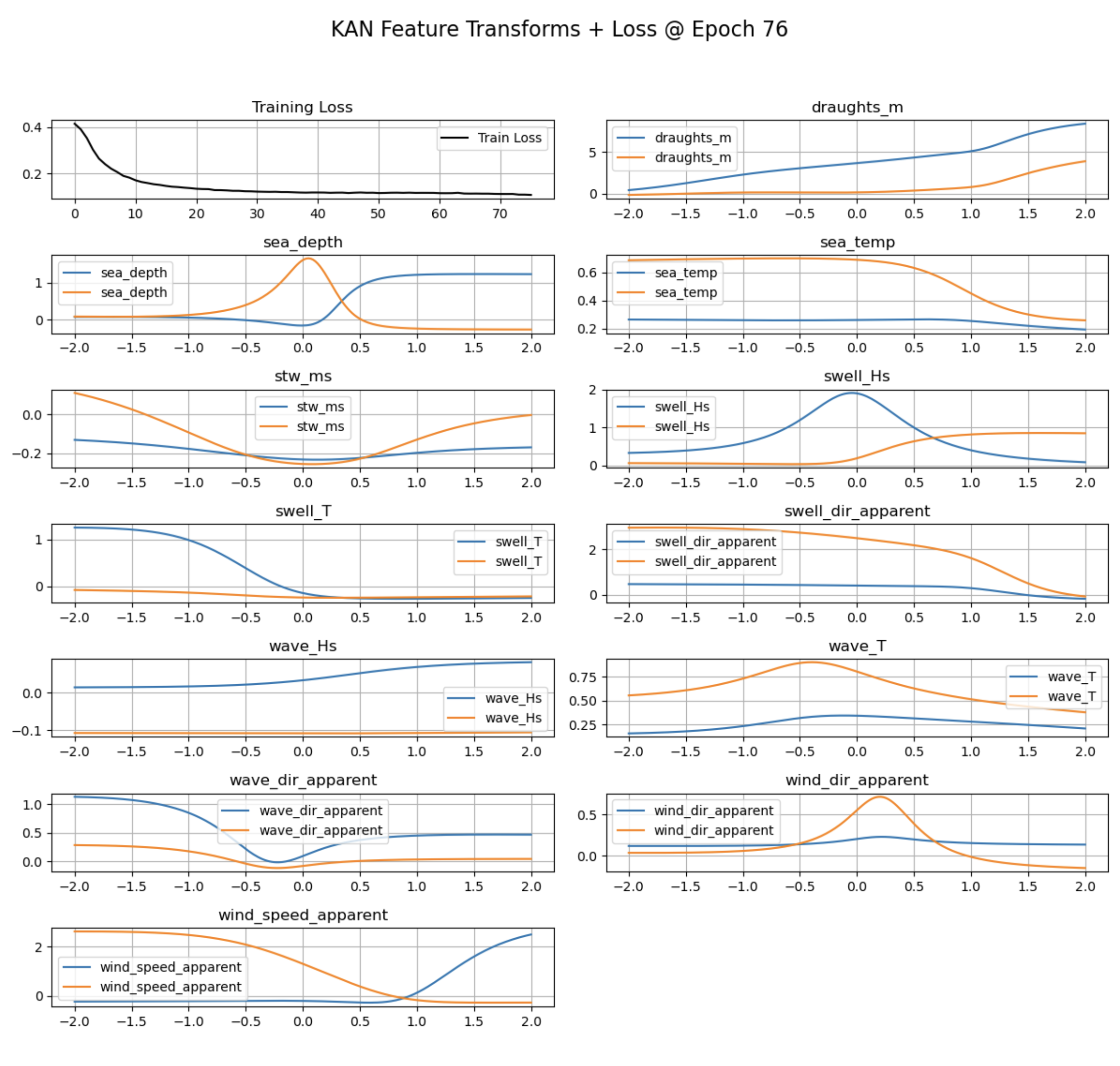}
\vspace{-2mm}
\caption{
Learned univariate feature transformation functions of the KAN model for shaft power prediction (epoch~76). Each subplot shows how an input feature is nonlinearly mapped before combination in the network. The transformations reflect physically meaningful patterns, including cubic-like dependence on ship speed through water (\texttt{stw\_ms}), cosine-like wind and wave directional effects, and localized nonlinear responses to swell and sea depth. The top-left subplot displays the training loss curve converging after $\sim$60~epochs.
}
\label{fig:kan_feature_transforms}
\vspace{-3mm}
\end{figure}

\subsection{Results for RQ3: Interpretability Insights for (PI-)KAN Models}\label{interpretability-insights}
To assess interpretability, we visualize the learned \emph{univariate} transformation functions (KAN splines) for each input feature (Figure~\ref{fig:kan_feature_transforms}). Unlike black-box neural networks, (PI-)KAN explicitly parameterizes and exposes feature-wise nonlinear mappings, enabling validation of model behavior against naval-architecture intuition.

The learned transformations recover physically plausible structures:
\begin{itemize}
  \item \textit{Speed and power: } The transformation for ship speed through water (\texttt{stw\_ms}) exhibits a cubic-like trend, consistent with the well-known cubic relationship between speed and resistance. Likewise, the rpm$\!\rightarrow\!$power mapping is near-cubic with low-rpm deviations (engine inefficiency) and mild high-rpm flattening (propeller ventilation/limits).
  
  \item \textit{Directional effects:} Apparent \texttt{wave\_dir\_apparent} and \texttt{wind\_dir\_apparent} show cosine-like patterns, capturing alignment between heading and environmental forcing (head vs.\ following seas/winds).

  \item \textit{Sea state and bathymetry:} \texttt{swell\_Hs} and \texttt{sea\_depth} display nonlinear peaks, highlighting regimes where hydrodynamic and added resistance grow more influential; wave-height effects are monotone beyond operationally relevant thresholds.
  
  \item \textit{Draft–power coupling:} The draft transformation yields a shallow U-shaped relation with an identifiable optimum, aligning with expected resistance–displacement trade-offs.
\end{itemize}

Training curves indicate stable convergence after \(\sim\)60 epochs. Taken together, these mappings explain \emph{why} PI-KAN achieves the strongest accuracy on energy-related targets (shaft power, fuel): gains arise from physically consistent behaviors rather than spurious correlations. In practice, the exposed splines provide actionable diagnostics (sanity checks for regime validity, early detection of sensor bias) and facilitate communication with operators through interpretable, physics-aligned responses.

\section{Discussion and Conclusion}
\label{discussion-and-conclusion}

 PI\textendash KAN advances vessel performance modelling by bridging physical laws and machine learning. It improves predictive accuracy, physical plausibility, and interpretability while offering insights into operational policies under challenging conditions. PI\textendash KAN consistently outperforms traditional baselines on \emph{shaft power} and \emph{fuel} across all vessels, while Polynomial regression remains competitive for \emph{shaft rpm}. Under rough weather, operational policy governs prediction error: adaptive slowdown reduces power/fuel variance and bias, whereas constant-speed strategies amplify error. 

\paragraph{Limitations.}
Despite strong vessel-wide performance, PI-KAN has limitations: (1) its additive, univariate design under-represents higher-order interactions (e.g., draft–trim–sea state coupling); (2) physics-regularization introduces hyperparameter sensitivity, and embedded laws assume quasi-steady conditions \cite{TaskarYumSteenPedersen2016}; (3) training costs are higher than classical baselines, complicating on-board deployment without pruning or distillation 
\cite{AlgarniAcarerAhmad2024}; and (4) KAN remains an evolving architecture \cite{faroughi2025scientific}, requiring further research for industrial adoption.

\paragraph{Challenges in Vessel-Wide Generalisation.}
Model robustness is affected by domain shifts between voyages (loading, fouling, weather), class imbalance in rare adverse conditions, and sensor biases \cite{DurlikMillerCembrowskaLechKrzemiskaothers2023}. Global physical constants may not fully capture vessel-specific deviations, and operational policies (e.g., maintaining a constant speed under high seas) interact with sea state to amplify variance and bias.

\paragraph{Future Directions.}
Future work will pursue \emph{selective interactions} to capture sparse multivariate effects while preserving interpretability~\cite{CaiLiWangLi2025}; \emph{uncertainty quantification} via Bayesian PI\textendash KAN for calibrated predictive intervals; \emph{continual calibration} to adapt online to drift and regime shifts~\cite{ZhangGaoCaoDongZouWangothers2022}; \emph{hybrid modeling} that couples Polynomial (rpm) with PI\textendash KAN (power/fuel) and employs model compression for edge deployment; and \emph{policy-aware bias correction} using voyage-specific bias heads to mitigate control-induced extremes.

\section*{Data and Code Availability}
The datasets and source code underlying this study are subject to contractual confidentiality and data-sharing restrictions with the industrial partner and therefore cannot be made publicly available.

\section*{Acknowledgment}
This work was supported by the Research Council of Norway through the GASS project (Grant No.\ 346603) and the eX3 infrastructure (Contract No.\ 270053). We used data from the E.U. Copernicus Marine Service Information\footnote{\url{https://doi.org/10.48670/moi-00016}} \footnote{\url{https://doi.org/10.48670/moi-00017}} \footnote{\url{https://doi.org/10.48670/moi-00021}} \footnote{\url{https://doi.org/10.48670/moi-00022}}

The authors thank Joachim Haga, Glenn Terje Lines, Dogan Altan, Akriti Sharma, and Tetyana Kholodna for their contributions to the empirical resistance formulas and the modeling formulation, as well as for valuable discussions on the methodology.

\bibliographystyle{IEEEtran}
\bibliography{ref}

\end{document}